\pdfoutput=1

\documentclass[11pt]{article}

\usepackage[final]{acl}

\usepackage{times}
\usepackage{latexsym}

\usepackage[T1]{fontenc}

\usepackage[utf8]{inputenc}

\usepackage{microtype}

\usepackage{inconsolata}

\usepackage{graphicx}
\usepackage{multirow}
\usepackage{pifont}
\usepackage{xcolor}
\usepackage{booktabs}
\usepackage{hyperref}
\usepackage{makecell}
\usepackage{amsmath}
\usepackage{amsfonts}
\usepackage{enumitem}
\usepackage{multicol}
\usepackage[ruled,vlined]{algorithm2e}
\usepackage{algpseudocode}
\usepackage{soul}
\usepackage[most]{tcolorbox}
\tcbuselibrary{skins}
\tcbuselibrary{breakable}

\definecolor{mintleaf}{RGB}{0, 184, 148}
\definecolor{dm-blue-500}{RGB}{0, 69, 177}
\definecolor{dm-purple-500}{RGB}{105,50,230}
\definecolor{mysilver}{RGB}{128,129,128}
\definecolor{my_green}{RGB}{0, 176, 80}
\definecolor{my_yellow}{RGB}{255,165,0}
\definecolor{my_red}{RGB}{255, 0, 0}
\definecolor{my_purple}{RGB}{126, 100, 158}
\definecolor{my_blue}{RGB}{49, 133, 155}
\definecolor{case_purple}{RGB}{160, 43, 147}
\definecolor{case_blue}{RGB}{15, 158, 213}

\newcommand{\cmark}{\textcolor{my_green}{\ding{51}}} 
\newcommand{\xmark}{\textcolor{my_red}{\ding{55}}} 

\newcommand{\method}{ETO}

\title{
\vspace{-2em}{\small \hfill ACL 2024 Main Conference}\\
\vspace*{.2in}
Trial and Error: Exploration-Based Trajectory Optimization\\ for LLM Agents
}

\author{Yifan Song$^{2,3}$~~Da Yin$^4$~~Xiang Yue$^5$~~Jie Huang$^6$~~Sujian Li$^{2,3}$~~Bill Yuchen Lin$^{1}$\\[10pt]
$^1$Allen Institute for AI \quad
$^2$School of Computer Science, Peking University \\
$^3$National Key Laboratory for Multimedia Information Processing, Peking University \\
$^4$UCLA \quad $^5$Ohio State University \quad $^6$UIUC \\[5pt]
{\small{\texttt{\{yfsong, lisujian\}@pku.edu.cn \quad yuchenl@allenai.org}}}
}

\begin{document}
\maketitle
\begin{abstract}


Large Language Models (LLMs) have become integral components in various autonomous agent systems.
In this study, we present an exploration-based trajectory optimization approach, referred to as \method{}. This learning method is designed to enhance the performance of open LLM agents. 
Contrary to previous studies that exclusively train on successful expert trajectories, our method allows agents to learn from their exploration failures. This leads to improved performance through an iterative optimization framework.
During the exploration phase, the agent interacts with the environment while completing given tasks, gathering failure trajectories to create contrastive trajectory pairs.
In the subsequent training phase, the agent utilizes these trajectory preference pairs to update its policy using contrastive learning methods like DPO~\citep{rafailov2023direct}.
This iterative cycle of exploration and training fosters continued improvement in the agents.
Our experiments on three complex tasks demonstrate that \method{} consistently surpasses baseline performance by a large margin.
Furthermore, an examination of task-solving efficiency and potential in scenarios lacking expert trajectory underscores the effectiveness of our approach.\footnote{Code \& Data: \href{https://github.com/Yifan-Song793/ETO}{https://github.com/Yifan-Song793/ETO}.}

\end{abstract}

\section{Introduction}

Large language models (LLMs) have demonstrated impressive capabilities in addressing complex interactive tasks through action planning for interactions with environments and tools~\citep{wang2023survey,xi2023rise}.
Systems using ChatGPT~\citep{chatgpt} and GPT-4~\citep{gpt4} as principal controllers have been developed for a range of applications. These include web browsing~\citep{deng2023mind2web,zhou2023webarena}, embodied tasks~\citep{yao2022react,lin2023swiftsage}, multi-modal reasoning~\citep{Lu2023ChameleonPC}, and complex question answering.
However, recent research suggests that open-source LLMs are considerably less effective than GPT-4 in constructing agents~\citep{liu2023agentbench,wang2023mint, mialon2023gaia}.

\begin{figure}[t]
    \centering
    \includegraphics[width=0.9\linewidth]{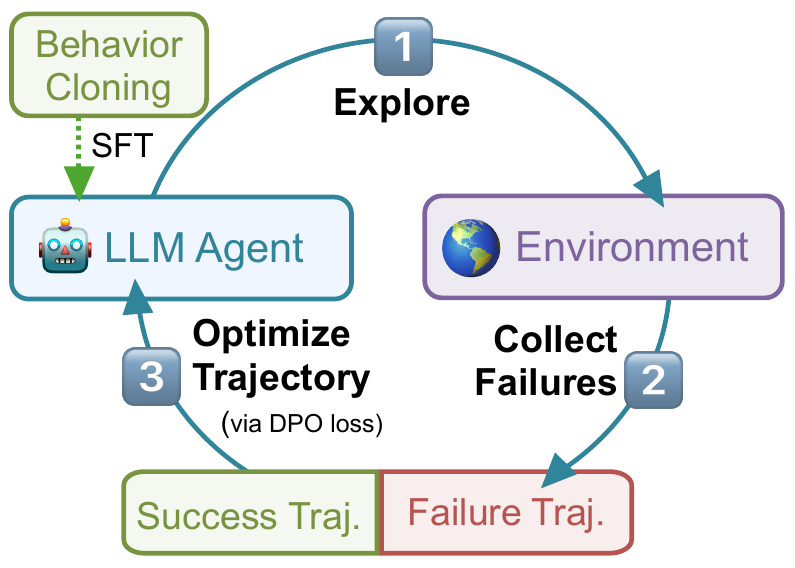}
    \caption{
    Exploration-based Trajectory Optimization (\method{}) allows an LLM agent to iteratively collect failure trajectories and update its policy by learning from contrastive failure-success trajectory pairs.
    }
    \label{fig:intro}
\end{figure}

The standard approach for constructing open LLM agents involves imitation learning, which fine-tunes LLMs based on expert trajectories.
Behavioral cloning (BC)~\citep{pomerleau1991efficient} is a simple yet effective imitation learning technique that derives a policy through supervised learning from observation-action pairs. 
Recent efforts~\citep{chen2023fireact,zeng2023agenttuning}, including Agent \textsc{Lumos}~\citep{yin2023lumos}, have explored the use of BC to develop open LLM agents by implementing supervised fine-tuning (SFT) on expert trajectories. 
These methods employ the teacher-forcing algorithm to train LLMs, enabling them to learn a policy for generating subsequent actions based on observations and past actions.  
However, these SFT methods, which rely entirely on expert demonstrations, may yield sub-optimal policies due to inadequate exploration of target environments, thereby limiting their generalizability.



The process of human learning not only involves observing successful demonstrations but also includes experiencing and exploring failures through trial-and-error interactions with the environment. 
Drawing inspiration from this, we propose a novel learning approach for LLM agents, which we call \textbf{Exploration-based Trajectory Optimization (\method{})}. 
Unlike previous methods that solely rely on successful trajectories, our approach capitalizes on the exploration failures of the current policy to enhance the learning process of an agent.

In particular, we first use SFT-based behavioral cloning to construct a base agent, as depicted in Figure~\ref{fig:intro}. 
During the \textit{exploration} phase, this base agent interacts with the target environment to undertake a set of given tasks and receive feedback from the environment.
We gather failed trajectories from the base agents and pair them with expert trajectories previously collected for those tasks.
In the subsequent \textit{training} phase, we apply the DPO loss~\citep{rafailov2023direct} to fine-tune the LLM policy with these contrastive trajectory pairs, thereby further improving the agent. 
The \method{} can be expanded to multiple rounds by collecting failure cases from previously ETO-tuned agents.


We assessed our approach using three representative datasets: \textit{WebShop}~\citep{yao2022webshop} for web navigation, \textit{ScienceWorld}~\citep{wang2022scienceworld} for simulated science experiments, and \textit{ALFWorld}~\citep{shridhar2020alfworld} for embodied household tasks.
Across these datasets, our \method{} consistently exceeded the performance of SFT behavioral cloning and other robust baselines by a significant margin, thereby demonstrating the effectiveness of learning from exploration failures.
Furthermore, our approach demonstrated an impressive performance improvement of $22\%$ over SFT on the challenging out-of-distribution test set in ScienceWorld, indicating its strong generalization capability.
Our analysis also highlighted the task-solving efficiency of our method, as it achieves higher rewards with fewer action steps.
In extreme scenarios where expert trajectories are not available, our \method{} still delivers impressive performance in a self-play mode, further emphasizing the potential of our approach.

In this paper, our contributions are as follows:
\begin{itemize}[leftmargin=*, nolistsep]
\setlength{\itemsep}{1mm}
    \item \textbf{Method.} We introduce exploration-based trajectory optimization, \method{}, a learning algorithm which iteratively collects failure trajectories and refines the agent policy via contrastive learning.
    \item \textbf{Evaluation.} Extensive experiments on three complex interactive tasks show that our method outperforms SFT behavioral cloning and other strong baselines by a large margin.
    \item \textbf{Analysis.} We conduct in-depth analysis to carefully validate the effectiveness of \method{} in multiple aspects, including out-of-distribution generalization, action efficiency, and its feasibility without the need for expert trajectories. 
\end{itemize}

\begin{figure*}[t]
    \centering
    \includegraphics[width=\linewidth]{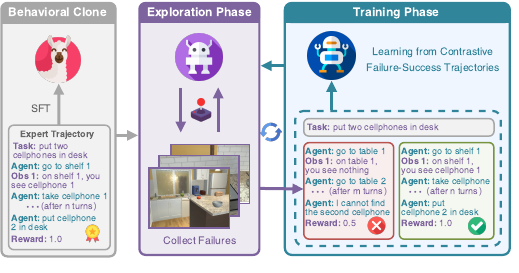}
    \caption{
    An illustrative overview of \textbf{Exploration-based Trajectory Optimization} (\method{}). Starting from a base LLM agent trained through behavioral cloning, our method allows the agent to iteratively collect failure trajectories and update its policy by continually learning from contrastive failure-success trajectory pairs.
    }
    \label{fig:main}
\end{figure*}

\section{Task Formulation}

The agent task with environment feedback can be formalized as a partially observable Markov decision process (POMDP) $\left(\mathcal{U}, \mathcal{S}, \mathcal{A}, \mathcal{O}, \mathcal{T}, \mathcal{R}\right)$ with instruction space $\mathcal{U}$, state space $\mathcal{S}$, action space $\mathcal{A}$, observation space $\mathcal{O}$, transition function $\mathcal{T}: \mathcal{S}\times\mathcal{A}\rightarrow\mathcal{S}$, and reward function $\mathcal{R}:\mathcal{S}\times\mathcal{A}\rightarrow\left[0,1\right]$.
Note that in our LLM-based agent scenario, $\mathcal{U}, \mathcal{A}, \mathcal{O}$ are subsets of natural language space.

Given a task instruction $u\in\mathcal{U}$, the LLM agent with parameter $\theta$ generates the action $a_1\sim \pi_\theta(\cdot|u) \in\mathcal{A}$ according to its policy $\pi_\theta$.
The action incurs a change in the latent state space $s_t\in\mathcal{S}$, and an execution feedback as observation $o_t\in\mathcal{O}$.
Then the agent generates the corresponding action in the $t+1$ step $a_{t+1}\sim \pi_\theta(\cdot|u,a_1,o_1,...,o_{t-1},a_t) \in\mathcal{A}$.
The interaction loop repeats until the task completes or exceeds the maximum steps, and the trajectory is denoted as:
\begin{align}
    &e=\left(u,a_1,o_1,...,o_{n-1},a_n\right) \sim \pi_\theta(e\vert u), \\
    &\pi_\theta(e\vert u)=\prod_{j=1}^n\pi_\theta(a_j|u,a_1,o_1,...,o_{j-1}),
\end{align}
where $n$ is the trajectory length.
Finally, the final reward $r(u,e)\in\left[0,1\right]$ is computed, with $1$ representing successful task completion.

\section{Method}

Our method, \method{}, starts by training a base agent through behavioral cloning. 
Based on the base agent, our framework continually enhances the policy from trial and error in an iterative manner.

\subsection{Behavioral Cloning}

Behavioral cloning (BC) has demonstrated promising results through supervised fine-tuning on the expert interaction trajectory data, serving as a solid starting point for building a powerful agent.
In this work, we employ ReAct-style~\citep{yao2022react} trajectory to conduct BC, which additionally generates Chain-of-Thought (CoT) rationales~\citep{wei2022chain} before each action.
Considering that the CoT and action are generated together in the ReAct framework, we use $a$ to represent the action with CoT for simplicity.

Given an expert trajectory dataset $\mathcal{D}=\left\{\left(u,e\right)^{\left(i\right)}\right\}_{i=1}^{\lvert \mathcal{D}\rvert}$, where $\lvert \mathcal{D}\rvert$ is the number of trajectories, we fine-tune an LLM on auto-regressive loss to get the base agent $\pi_{\mathrm{base}}$:
\begin{equation}
\label{eq:sft}
    \mathcal{L}_{\mathrm{SFT}} (\pi_\theta) = -\mathbb{E}_{e\sim\mathcal{D}}\left[\pi_\theta(e\vert u) \right],
\end{equation}
where $e=\left(u,a_1,o_1,...o_{n-1},a_n\right)\sim\mathcal{D}$ is an expert interaction trajectory.

Since $\pi_\theta(e\vert u)=\prod_{j=1}^n\pi_\theta(a_j|u,...,o_{j-1})$, in practice, we first concatenate the instruction, actions and observations in trajectory $e$ as a text sequence $t$:\looseness=-1
\begin{equation}
\begin{aligned}
    t&=\mathrm{concat}(u,a_1,o_1,...,o_{n-1},a_n) \\
    &=\left(t_1,t_2,...,t_l\right),
\end{aligned}
\end{equation}
where $t_k$ is the $k$-th token in the result sequence.
Then the probability of the trajectory in Eq.~\eqref{eq:sft} can be obtained by directly computing the probability of actions with tokens in task description and observations masked:
\begin{equation}
    \pi_\theta(e\vert u)=-\sum_k\log \pi_\theta(t_k|t_{<k})\times \mathbf{1}(t_k\in A),
\end{equation}
where $\mathbf{1}(t_k\in A)$ is an indicator function about whether $t_k$ is a token belonging to actions produced by the agent.

\begin{algorithm*}[t!]
\DontPrintSemicolon
\KwIn{$\mathcal{D} = \left\{\left(u,a_1,o_1,...o_{n-1},a_n\right)^{\left(i\right)}\right\}$: expert trajectory dataset for behavioral cloning, $T_1$: number of behavioral cloning steps, $I$: number of iterations for \method{}, $T_2$: number of steps in training phase, $\pi_\theta$: initial LLM policy.}
\KwOut{Final policy $\pi_{\theta}$}
~\textcolor{dm-purple-500}{\texttt{// Behavioral cloning}} \;
\For{$i= 1$ \KwTo $T_1$}{
    Optimize $\theta$ on BC objective: $\mathcal{L}_{\mathrm{SFT}} (\pi_\theta) = -\mathbb{E}_{e\sim\mathcal{D}}\left[\pi_\theta(e\vert u) \right]$
}
~\textcolor{dm-blue-500}{\texttt{// Iteratively learning from exploration failures}}\;
\For{$i= 1$ \KwTo \em $I$}{
    $\pi_\mathrm{base}=\pi_\theta$; $\pi_\mathrm{ref}=\pi_\theta$ \;
    Get base agent trajectories on $\mathcal{D}$: $\hat{e}=(u,\hat{a}_1,\hat{o}_1,...,\hat{o}_{m-1},\hat{a}_m) \sim \pi_\mathrm{base}(e\vert u)$ \;
    Compare rewards of $\hat{e}$ with expert trajectory $e$ to get the failure-success pair: $e_w\succ e_l\ \vert\ u$ \;
    Construct contrastive trajectory dataset: $\mathcal{D}_p=\left\{ \left(u,e_w,e_l \right)^{(i)} \right\}$ \;
    \For{$j= 1$ \KwTo \em $T_2$}{
        Optimize $\theta$ on trajectory contrastive objective: $\mathcal{L}_\mathrm{DPO}(\pi_\theta;\pi_\mathrm{ref})=-\mathbb{E}_{(u,e_w,e_l)\sim \mathcal{D}_p}\Bigg[\log \sigma \Big(\beta \log\frac{\pi_\theta(e_w\vert u)}{\pi_\theta(e_l\vert u)} - \beta\log\frac{\pi_\mathrm{ref}(e_w\vert u)}{\pi_\mathrm{ref}(e_l\vert u)}\Big)\Bigg]$
    }
}
\Return{$\pi_{\theta}$}
\caption{\textbf{\method{}: Exploration-based Trajectory Optimization for LLM Agents}}
\label{algo:left}
\end{algorithm*}

\subsection{Learning From Exploration Failures}

Behavioral cloning exclusively depends on expert trajectories and lacks the ability to explore the environment, leading to sub-optimal policies. 
To train a more powerful agent, it is important for the model to also explore failure trajectories. 
To achieve this, a viable approach is reinforcement learning, which empowers agents to actively explore the environment to get rewards and refine the policy through trial and error~\citep{ouyang2022training}:
\begin{equation}
\begin{split}
\label{eq:rlhf}
    \max_{\pi_\theta}\mathbb{E}_{u\sim\mathcal{D},e\sim\pi_\theta\left(e\vert u\right)} & \left[r(u,e)\right]- \\
    \beta \mathbb{D}_{\mathrm{KL}} & \left[\pi_\theta (e\vert u)\ \vert\vert\ \pi_\mathrm{ref}(e\vert u)\right],
\end{split}
\end{equation}
where the KL term with weighting parameter $\beta$ controls the deviation from the base reference policy $\pi_\mathrm{ref}$, \textit{i.e.}, the base agent $\pi_\mathrm{base}$.
In practice, the agent to be trained $\pi_\theta$ is also initialized to $\pi_\mathrm{base}$.
Then the optimization problem in Eq.~\eqref{eq:rlhf} can be solved via RL methods such as PPO~\citep{schulman2017proximal,ouyang2022training}.

However, directly applying online RL on LLM agents will present practical challenges such as instability and low efficiency~\citep{shen2023large,rafailov2023direct}.
Therefore, we instead design an iterative offline learning framework and train the agent with contrastive trajectory data.
As shown in Figure \ref{fig:main}, the training process can be formulated in an iterative exploration-training loop.
In the exploration phase of \method{}, the agent explores the environment to collect failure trajectories.
During the training phase, the agent learns the contrastive information from the ``failure-success'' trajectory pairs to update the policy.

\paragraph{Exploration Phase}

In this phase, the base agent $\pi_{\mathrm{base}}$ explores the environment to get the trajectories on the instructions of training data for BC:
\begin{equation}
    \hat{e}=(u,\hat{a}_1,\hat{o}_1,...,\hat{o}_{m-1},\hat{a}_m) \sim \pi_\mathrm{base}(e\vert u).
\end{equation}
The environments then return a reward $\hat{r}$ corresponding to the trajectory $\hat{e}$.

Then we construct failure-success trajectory pairs, denote as $e_w\succ e_l\ \vert\ u$, based on the final rewards.
Here, $e_w$ and $e_l$ represent the trajectories with higher and lower rewards, chosen from the expert trajectory $e$ and agent-generated trajectory $\hat{e}$ respectively.
Note that we only collect pairs where two trajectories have different rewards.
If both $\hat{e}$ and $e$ successfully complete the task, we simply discard the pair.
Finally, we get the contrastive trajectory dataset $\mathcal{D}_p=\left\{ \left(u,e_w,e_l \right)^{(i)} \right\}_{i=1}^{\lvert \mathcal{D}_p \rvert}$.

\paragraph{Training Phase}
\label{sec:training}

In this phase, the agent policy is updated by modeling the contrastive failure-success information in the trajectory pair data.

Given trajectory pair $e_w\succ e_l\ \vert\ u$, the failure-success relation can be modeled via Bradley-Terry (BT)~\citep{bradley1952rank} model:
\begin{equation}
\label{eq:bt}
    p(e_w\succ e_l\vert u)=\frac{\exp\left(r(u,e_w)\right)}{\exp\left(r(u,e_w)\right) + \exp\left(r(u,e_l)\right)}.
\end{equation}
Under the optimal policy $\pi_r(e\vert u)$ of Eq.~\eqref{eq:rlhf}, the reward function in the environment can be written as~\citep{peng2019advantage,rafailov2023direct}:
\begin{equation}
\label{eq:reward}
    r(u,e)=\beta \log\frac{\pi_r(e\vert u)}{\pi_\mathrm{ref}(e\vert u)}+\beta \log Z(x),
\end{equation}
where $Z(u)=\sum_e \pi_\mathrm{ref}(e\vert u)\exp\left(\frac{1}{\beta}r(u,e)\right)$ is the partition function.
Substitute Eq.~\eqref{eq:reward} into Eq.~\eqref{eq:bt} to get the BT model over policy:
\begin{equation}
\begin{split}
    p(e_w & \succ e_l\vert u)= \\
    &\sigma \left(\beta \log\frac{\pi_\theta(e_w\vert u)}{\pi_\theta(e_l\vert u)} - \beta\log\frac{\pi_\mathrm{ref}(e_w\vert u)}{\pi_\mathrm{ref}(e_l\vert u)}\right),
\end{split}
\end{equation}
where $\sigma$ is the sigmoid function.
Then the optimal policy $\pi_\theta$ can be achieved by applying maximum likelihood:
\begin{equation}
\label{eq:dpo}
\resizebox{1.0\hsize}{!}{$
\begin{aligned}
    \mathcal{L}_\mathrm{DPO}&(\pi_\theta;\pi_\mathrm{ref})=\\
    &-\mathbb{E}_{(u,e_w,e_l)\sim \mathcal{D}_p}\Bigg[\log \sigma \Big(\beta \log\frac{\pi_\theta(e_w\vert u)}{\pi_\theta(e_l\vert u)} - \beta\log\frac{\pi_\mathrm{ref}(e_w\vert u)}{\pi_\mathrm{ref}(e_l\vert u)}\Big)\Bigg].
\end{aligned}
$}
\end{equation}

This optimization objective aims to increase the likelihood of the success trajectories $e_w$ and decrease the likelihood of failure trajectories $e_l$, with a constraint term to maintain the basic agent capabilities.
Moreover, as a reformulation of RL objective Eq.~\eqref{eq:rlhf}, Eq.~\eqref{eq:dpo} is directly maximizing the final reward while avoiding the need to perform RL optimization.

\paragraph{Iteration}

To further improve the agent's performance, \method{} adopts an iterative exploration-training manner.
After the training phase, the agent policy can be used to gather new failure cases and create contrastive trajectory pairs.
These new data are then used to further enhance the agent through trajectory contrastive learning.
The complete learning process of \method{} is shown in Algorithm~\ref{algo:left}.

\section{Experiments}

In this section, we conduct extensive experiments to validate the effectiveness of \method{}.
Our method demonstrates superior performance compared to baselines across three datasets, and it exhibits enhanced advantages when dealing with out-of-domain unseen tasks.
The analysis further showcases the efficiency of our method.
Furthermore, our method also holds promise in scenarios where expert trajectories are unavailable.

\subsection{Experimental Settings}

\paragraph{Datasets}

We conduct experiments on three representative agent datasets, WebShop~\citep{yao2022webshop} for web navigation, ScienceWorld~\citep{wang2022scienceworld} for embodied science experiments, and ALFWorld~\citep{shridhar2020alfworld} for embodied house holding tasks.
Both WebShop and ScienceWorld environments provide dense final rewards ranging from 0 to 1, while ALFWorld only provides binary rewards indicating whether the task is completed.
All three environments can be formally described as partially observable Markov decision processes. 
For details of the datasets and the expert trajectory collection process, please refer to Appendix~\ref{app:datasets}.

The statistical information of our datasets is presented in Table \ref{tab:dataset}.
It is important to mention that, in addition to the in-distribution test sets, both ScienceWorld and ALFWorld contain test sets that include out-of-distribution unseen variations. 
These additional test sets allow us to assess the generalization capabilities of different agents.

\begin{table}[t]
\centering
\tabcolsep=3pt
\resizebox{\linewidth}{!}{
\begin{tabular}{lcccc}
\toprule
\textbf{Dataset} & \#Train & \#Test-Seen & \#Test-Unseen & \#Turns \\
\midrule
WebShop & 1938 & 200 & - & 4.9 \\
ScienceWorld & 1483 & 194 & 241 & 14.4 \\
ALFWorld & 3321 & 140 & 134 & 10.1 \\
\bottomrule
\end{tabular}
}
\caption{
Statistics of datasets. ``Test-Seen'' and ``Test-Unseen'' are test set with seen and unseen scenarios respectively. ``\#Turns'' denotes the average number of interaction turns for the expert trajectories.
}
\label{tab:dataset}
\end{table}

\paragraph{Training Setup}

We mainly use Llama-2-7B-Chat~\citep{touvron2023llama} as the base model for building LLM agents. 
To provide more comprehensive results, we also conduct experiments on Llama-2-13B-Chat and Mistral-7B~\citep{jiang2023mistral}.
We utilize the AdamW optimizer~\citep{loshchilov2017decoupled}.
For the SFT phase, the batch size is 64 and the learning rate is set to 1e-5 with $3\%$ warm up and a cosine scheduler.
Then the base agent will explore once for each instance in the training set to collect failure trajectories.
For the training phase of \method{}, the batch size is 32 and the learning rate is set to 1e-6.
The $\beta$ in DPO loss is set to $0.1$ for WebShop and ScienceWorld, $0.5$ for ALFWorld.
The learning epochs of SFT phase and training phase in \method{} are set to 3.
The number of iterations of \method{} is set to 2 for WebShop and ScienceWorld, 1 for ALFWorld.
All experiments are conducted on 8 NVIDIA A100 80G GPUs.

\begin{table*}[t]
\vspace{-1em}
\centering
\tabcolsep=15pt
\resizebox{1\linewidth}{!}{
\begin{tabular}{@{}lccccc@{}}
\toprule
\multirow{2}{*}{\textbf{Method}} & \multirow{2}{*}{\textbf{WebShop}} & \multicolumn{2}{c}{\textbf{ScienceWorld}} & \multicolumn{2}{c}{\textbf{ALFWorld}} \\
\cmidrule(l){3-4} \cmidrule(l){5-6}
& & Seen & Unseen & Seen & Unseen \\
\midrule
GPT-4 & 63.2 & 64.8 & 64.4 & 42.9 & 38.1 \\
GPT-3.5-Turbo & 62.4 & 16.5 & 13.0 & 7.9 & 10.5 \\
\midrule
Llama-2-7B-Chat & 17.9 & 3.8 & 3.1 & 0.0 & 0.0 \\
Llama-2-7B-Chat + SFT & 63.1 & 67.4 & 53.0 & 60.0 & 67.2 \\
Llama-2-7B-Chat + Best-of-N & 63.8 & 70.2 & 57.6 & 62.1 & 69.4 \\
Llama-2-7B-Chat + RFT & 63.6 & 71.6 & 54.3 & 62.9 & 66.4 \\
Llama-2-7B-Chat + PPO & 64.2 & 59.4 & 51.7 & 22.1 & 29.1 \\
\midrule
Llama-2-7B-Chat + \method{} (ours) & \textbf{67.4} & \textbf{73.8} & \textbf{65.0} & \textbf{68.6} & \textbf{72.4} \\
\bottomrule
\end{tabular}
}
\caption{
The average reward of different methods on three agent datasets. ``Seen'' denotes the held-out test set with task types seen during training, while ``Unseen'' refers to the test set with critical unseen task variations.
}
\label{tab:main}
\end{table*}

\paragraph{Baselines}

We compare \method{} with SFT behavioral cloning and other post-imitation baseline methods.
1) SFT~\citep{chen2023fireact,zeng2023agenttuning} conducts behavioral cloning on expert trajectories, which is the base agent for \method{} and other baselines.
2) Best-of-N sampling employs SFT base agent and selects the trajectory with the best reward within N samplings. Here we set N to 10.
3) RFT (Rejection sampling Fine-Tuning)~\citep{yuan2023scaling} is a strong baseline which adds the success trajectories to the expert trajectory dataset and trains the agent on new augmented trajectories.
4) PPO (Proximal Policy Optimization)~\citep{schulman2017proximal} is an RL method directly optimizing the SFT agents to maximize the final task reward.
We also include GPT-3.5-Turbo~\citep{chatgpt}, GPT-4~\citep{gpt4}, and untuned Llama-2-7B-Chat for comparison.

\paragraph{Evaluation}
All methods are evaluated using the ReAct-style interaction format~\citep{yao2022react}, with CoT rationale generated before the action.
See Appendix \ref{app:prompt} for the detailed prompts.
We add 1-shot in-context example in the instruction prompt for each task.
The decoding temperature of the LLMs is set to be $0.0$ for deterministic generation, except for Best-of-N method.
We mainly employ \textbf{Average Reward} as the metric, which represents the average reward of all task instances in the test set.
We also report \textbf{Success Rate} in Appendix~\ref{app:sr} for reference.

\subsection{Results}

Table \ref{tab:main} presents the performance comparison of \method{} and baselines on three agent datasets.
As shown, \method{} demonstrates a significant improvement over SFT imitation learning, leading to an average reward increase of  $8\%$ and $9.5\%$ for WebShop and ScienceWorld.
Our method also outperforms all other baselines on all datasets.
On WebShop dataset, \method{} even outperforms GPT-4 on the average reward, showing the extraordinary performance of our method.
Although the RFT method also exhibits improvement compared to SFT, its performance remains constrained as it is an augmented version of behavioral cloning and solely learns from success trajectories.
This comparison indicates the comparison between failure and expert trajectories is essential to improve the performance of the agent.
Meanwhile, though PPO gains improved performance on WebShop, it struggles to achieve satisfactory results on the other two datasets due to the inherent instability in RL optimization procedures, particularly on ALFWorld dataset which only provides binary final rewards.
In Appendix \ref{app:case}, we present case studies to show the task-solving trajectories of our method.

Notably, our approach showcases enhanced advantages in out-of-domain unseen scenarios, achieving an impressive performance boost of $20\%$ on ScienceWorld-Unseen.
Moreover, \method{} exhibits strong effectiveness on the unseen scenarios in ALFWorld and outperforms the RFT and PPO baselines, both of which suffer from performance degradation.
These results underscore that learning by trial and error can further enhance the agent's generalization capabilities, particularly in out-of-distribution unseen scenarios.

\paragraph{Results on Different LLMs}

\begin{table}[t]
\centering
\resizebox{\linewidth}{!}{
\begin{tabular}{lcccc}
\toprule
\multirow{2}{*}{\textbf{Base LLM}} & \multirow{2}{*}{\textbf{Method}} & \multirow{2}{*}{\textbf{WebShop}} & \multicolumn{2}{c}{\textbf{ScienceWorld}} \\
\cmidrule(l){4-5}
& & & Seen & Unseen \\
\midrule
\multirow{2}{*}{Llama-2-13B} & SFT & 66.3 & 68.1 & 57.6 \\
& \method{} & 70.7 & 71.4 & 68.6 \\
\midrule
\multirow{2}{*}{Mistral-7B} & SFT & 60.1 & 63.8 & 52.2 \\
& \method{} & 66.2 & 68.5 & 62.5 \\
\bottomrule
\end{tabular}
}
\caption{
The average reward of different base LLMs on WebShop and ScienceWorld.
}
\label{tab:llms}
\end{table}

To further demonstrate the effectiveness of our method, we present the results based on other base LLMs, including Llama-2-13B-Chat and Mistral-7B.
Table \ref{tab:llms} showcases the consistent improvement in agent performance achieved by \method{} across different LLMs.
Notably, when compared to Llama-2-7B, the 13B model displays a relatively smaller performance gain on both datasets, suggesting that our method can provide greater benefits to weaker agents.
Despite Mistral-7B is a more powerful LLM than Llama-2-13B, it falls short of Llama-2-7B after either SFT or \method{}.
This finding indicates that there is not a strong correlation between the basic LLM capabilities and the agent capabilities.

\paragraph{Analysis on Efficiency}

\begin{figure}[t]
    \centering
    \includegraphics[width=\linewidth]{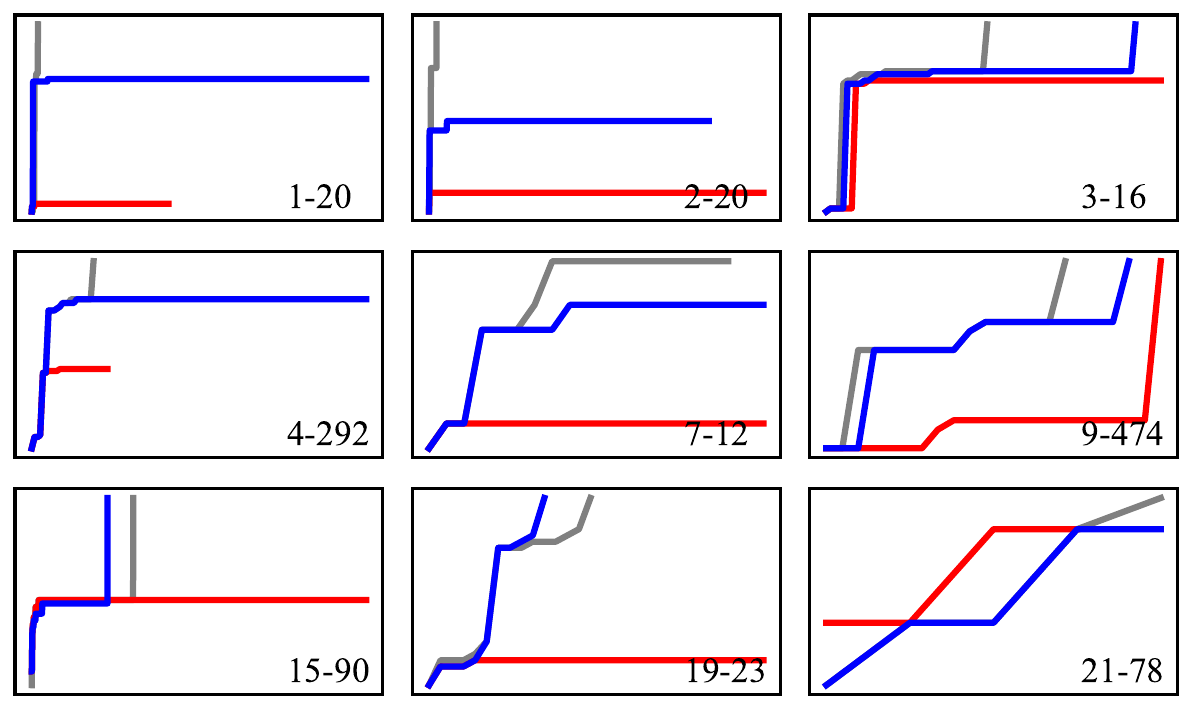}
    \caption{
    Cases of ScienceWorld reward trajectory for \textcolor{blue}{\method{}}, \textcolor{red}{SFT Base Agent} and \textcolor{mysilver}{Oracle}.
    $X$: time steps $(0\rightarrow T)$; $Y$: scores $(0\rightarrow 100)$.
    Task IDs are shown at the bottom-right. Best viewed in color.
    }
    \label{fig:efficiency}
\end{figure}

We evaluate the task-solving efficiency of agents in ScienceWorld environment, which provides fine-grained subgoals for each task.
The reward of a task is updated upon the accomplishment of a subgoal.
By assessing the agent's ability to achieve higher rewards within fewer action steps, we can determine its efficiency.
Figure \ref{fig:efficiency} showcases the score trajectories of ScienceWorld-Seen test set, comparing \method{} with the SFT base agent, and the oracle agent.
As depicted, \method{} can reach higher rewards in fewer action steps than the SFT base agent.
Interestingly, in certain cases like 15-90 and 19-23, our method outperforms even the oracle agent, reaching a score of 100 earlier.
These results demonstrate that by learning from failure trajectories, our method also acquires a more powerful task-solving efficiency.

\subsection{Ablation of Iterations}

\begin{figure}[t]
    \centering
    \includegraphics[width=\linewidth]{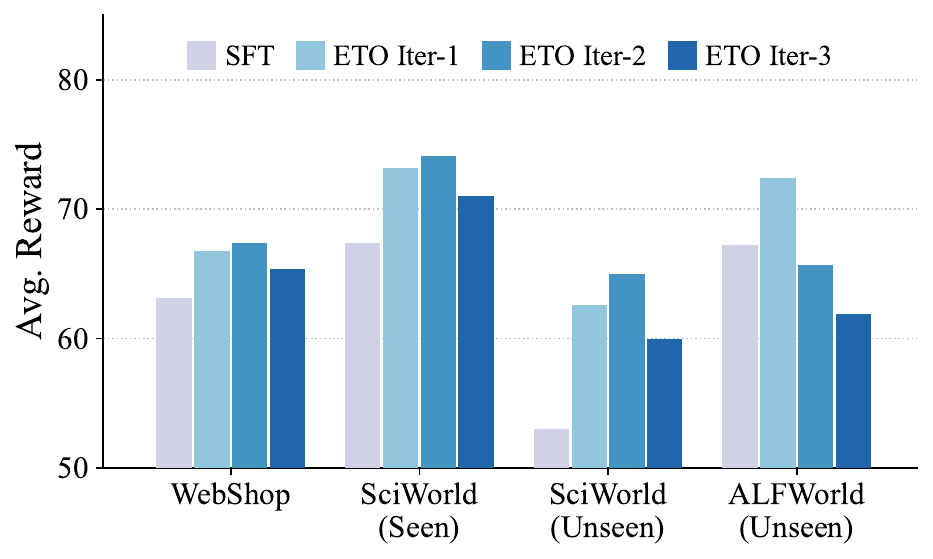}
    \caption{
    \method{} performance on multiple iterations.
    }
    \label{fig:iter}
\end{figure}

In this section, we present a study on the impact of iteration numbers in \method{}.
The results are shown in Figure \ref{fig:iter}.
As depicted, \method{} demonstrates the ability to enhance the performance of agents in the first two iterations on both the WebShop and ScienceWorld datasets.
However, further increasing the iterations does not lead to continuous improvement.
Instead, the performance starts to decline after the third iteration.
Regarding the ALFWorld dataset, only the first iteration of \method{} shows an improvement.
Surprisingly, the performance on the second and third iterations even falls behind that of the SFT base agent.

To explain this, it is important to note that the learning process of ETO relies on a fixed expert trajectory set, and the exploration phase of the agent is performed on the same training set.
Consequently, the diversity and quantity of failure-success contrastive trajectory data are constrained.
Initially, the policy can be improved by learning from past mistakes.
However, the model gets overfitting on the contrastive information in subsequent iterations, resulting in a decline in performance.
In the case of ALFWorld, the coarse-grained binary rewards further hinder the agent from getting improvement from iterative training.
As a potential solution, future work could explore the incorporation of GPT-4 to dynamically construct more diverse contrastive trajectory data.

\subsection{Strategy of Contrastive Data Construction}

In this section, we delve deeper into the contrastive trajectory pair construction strategy used in our method.
In Section \ref{sec:training}, we directly learn from failure-success trajectory pairs (Eq.~\eqref{eq:dpo}), referred to \textit{trajectory-wise contrastive}.
Alternatively, inspired by previous work~\citep{lightman2023let}, we introduce a fine-grained variation of \method{} that captures \textit{step-wise contrastive} information by comparing ``good-bad'' action pairs.
To achieve this, we use the expert trajectory to conduct teacher forcing for first $t-1$ steps, and then have the agent predict the action of $t$-th step.
Then the quality of $t$-th action is determined by the final rewards.
We also implement a \textit{mixture} variation by combining the above two strategies.
For further details regarding the step-wise variation of \method{}, please refer to Appendix \ref{app:action}.

\begin{table}[t]
\centering
\resizebox{0.95\linewidth}{!}{
\begin{tabular}{llccc}
\toprule
\textbf{Method} & \textbf{Level} & \textbf{lr} & $\beta$ & \textbf{Avg. Reward} \\
\midrule
SFT & - & - & - & 63.1 \\
\midrule
\multirow{4}{*}{\method{}} & Trajectory & 1e-6 & 0.1 & \textbf{67.4} \\
& Step & 1e-6 & 0.1 & 8.3 \\
& Step & 1e-7 & 0.5 & 62.8 \\
& Mixture & 1e-6 & 0.1 & 64.3 \\
\bottomrule
\end{tabular}
}
\caption{
The average reward on WebShop of agents trained on different level contrastive data.
}
\label{tab:preference}
\end{table}

The comparison of different methods is presented in Table \ref{tab:preference}.
As the results demonstrate, trajectory-wise contrastive yields the best performance.
On the other hand, we observed that step-wise contrastive modeling tends to be less stable, necessitating a lower learning rate and a higher constraint parameter $\beta$ to maintain the basic capabilities of the agent.
This instability may be attributed to the inaccurate estimation of the action quality, as we simply utilize the final rewards to construct step-wise contrastive pairs.
Moreover, the performance of mixture strategy also falls short compared to trajectory-level contrastive.

\subsection{Self-Play w/o Expert Trajectory}

In this section, we explore a challenging scenario where no expert trajectory is available.
In such cases, the agent is compelled to explore the environment and depend on self-play to enhance its capabilities.
To achieve this, we eliminate the behavioral cloning phase of \method{} and allow the LLM agent to explore the environments using a decoding temperature of $1.0$.
Subsequently, we compare different trajectories associated with the same instruction based on their final rewards, creating trajectory preference data.
Finally, the agent is trained exclusively on the preference data generated by itself.

On WebShop, the untuned Llama-2-7B-Chat achieves relatively high rewards.
Thus, we use this dataset to conduct the experiment.
We also employ rejection sampling fine-tuning (RFT) as a baseline.
The results in Table \ref{tab:cold_start} show that \method{} alone does not improve performance without behavioral cloning.
In contrast, RFT shows promising ability to enhance the agent's capabilities without relying on expert trajectories.
However, when combining RFT with \method{}, we observe a further enhancement in the agent's performance.
These findings suggest that in scenarios without expert trajectories, it may be beneficial to first employ RFT and then allow the agent to learn from exploration failures.
These results further highlight the potential of our method when expert trajectories are unavailable.

\begin{table}[t]
\centering
\tabcolsep=14pt
\resizebox{\linewidth}{!}{
\begin{tabular}{lcc}
\toprule
\textbf{Method} & \textbf{w/ BC?} & \textbf{Avg. Reward} \\
\midrule
SFT & \cmark & 63.1 \\
RFT & \cmark & 63.6 \\
\method{} & \cmark & 67.4 \\
\midrule
Llama-2-7B-Chat$^\dag$ & \xmark & 17.9 \\
RFT  & \xmark & 48.4 \\
\method{} & \xmark & 12.5 \\
RFT+\method{} & \xmark & \textbf{51.2} \\
\bottomrule
\end{tabular}
}
\caption{
Performance of self-play without behavioral cloning from expert trajectories.
Methods in the upper part are implemented upon a BC base agent, while the methods in the lower part directly use the untuned Llama as the starting point.
$\dag$ means directly prompting an untuned Llama-2-7B-Chat.
}
\label{tab:cold_start}
\end{table}

\section{Related Work}

\paragraph{Imitation Learning} 
Imitation learning is a learning paradigm where an agent learns a policy by mimicking expert demonstrations~\citep{hussein2017imitation,fang2019survey}.
A prevalent approach in imitation learning is behavioral cloning (BC)~\citep{pomerleau1991efficient}, which utilizes expert trajectories to learn a direct mapping from states to actions.
There are various methods to mitigate the limitations of BC~\citep{ross2011reduction,ross2014reinforcement}.
Our method, \method{}, shares a similar spirit with DAgger~\citep{ross2011reduction}, an approach used to enhance the agent's performance by learning from failure cases.
However, unlike DAgger which gathers additional expert trajectories on agent-failed cases, \method{} improves the policy through learning from contrastive trajectory pairs.

\paragraph{LLM Agents}
With the various emergent abilities of LLMs, researchers have explored building agent systems based on LLMs~\citep{xi2023rise}.
Recent projects such as AutoGPT~\citep{autogpt}, BabyAGI~\citep{babyagi}, and RestGPT~\citep{song2023restgpt} have employed LLMs as core controllers, building powerful agent frameworks capable of solving realistic tasks.
While GPTs have shown strong agent intelligence, open-source LLMs still lag far behind~\citep{liu2023agentbench,wang2023mint}.
To bridge this gap, recent studies, including FireAct~\citep{chen2023fireact}, AgentTuning~\citep{zeng2023agenttuning}, and Lumos~\citep{yin2023lumos}, construct expert trajectory data from teacher agents (\textit{e.g.}, GPT-4) and perform BC on open-source LLMs.
Taking a step further, \citet{aksitov2023rest} refine the agent through iterative BC on success trajectories generated by the previous policy.
Concurrently with our work, \citet{yang2023embodied} use the DAgger framework~\citep{ross2011reduction} and also employ DPO loss to develop embodied multi-modal agents.

\begin{table}[t]
\centering
\tabcolsep=4pt
\resizebox{\linewidth}{!}{
\begin{tabular}{lcccc}
\toprule
\textbf{Method} & \textbf{Exploration} & \textbf{Reward} & \textbf{Efficiency} & \textbf{Robustness} \\
\midrule
SFT & \xmark & \xmark & \cmark & \cmark \\
RFT & \cmark & \xmark & \cmark & \cmark \\
Online RL & \cmark & \cmark & \xmark & \xmark \\
Offline RL & \xmark & \cmark & \cmark & \xmark \\
\midrule
\method{} & \cmark & \cmark & \cmark & \cmark \\
\bottomrule
\end{tabular}
}
\caption{
\method{} vs alternatives: \method{} can leverage exploration failures to optimize the policy with high computational efficiency and robustness.
}
\label{tab:compare}
\end{table}

\paragraph{LLM Policy Learning}
Learning from preference has shown promise for learning an enhanced LLM policy, particularly in LLM alignment research.
Reinforcement Learning from Human Feedback (RLHF) is a method that learns a reward model and then utilizes proximal policy optimization to update the policy model~\citep{christiano2017deep,ouyang2022training}.
Despite its attractive advantages, RLHF presents limitations regarding training efficiency and instability.
To address these issues, \citet{rafailov2023direct} reformulate the optimization objective of RLHF, introducing the DPO loss to directly model preferences.
Similar to our work, ReST~\citep{gulcehre2023reinforced} iteratively generates new samples from the current policy and refines the policy using offline RL methods.
Recent studies have explored the application of LLM policy learning in other domains~\citep{lightman2023let,wang2023making}.
For example, \citet{wang2023math} train a step-wise reward model to improve the performance of LLMs in mathematical reasoning.
The comparison of \method{} with several alternatives in Table~\ref{tab:compare} highlights the strength of our method in LLM agent policy learning.

\section{Conclusion}

In this work, we present \method{}, a method aimed at enhancing the capabilities of LLM agents.
Our approach allows the agent to learn by trial and error, thereby improving the performance of the base agent acquired through behavioral cloning.
\method{} uses an exploration-training iteration framework.
During the exploration phase, the agent explores the environment, gathering failure trajectories and constructing trajectory preference pairs.
Subsequently, in the training phase, the agent learns from the preference information using DPO loss.
This iterative process of exploration and training enables further improvement in the agent's performance.
Extensive experiments on three agent datasets demonstrate our method outperforms behavioral cloning and strong baselines by a large margin.
Moreover, our method exhibits remarkable efficiency and shows great potential in scenarios where expert trajectories are unavailable.

\section*{Limitations}

Our method, \method{}, demonstrates effective learning of powerful LLM agents through trial and error.
However, it is important to acknowledge several limitations of this work.
1) \method{} simplifies the comparison of failure-success trajectories by assuming that the agent generates wrong actions right from the beginning.
However, in realistic cases, the agent may start executing incorrect actions from some intermediate step.
If we can identify when the agent makes a bad action (\textit{e.g.}, $\hat{a}_3$ at $3$-th step), we should then collect the expert trajectory for the remaining actions $a_{t>3}$.
Unfortunately, most current environments do not contain such information, making it challenging to conduct action-wise or process-level reward modeling.
A potential solution is to employ GPT-4 to identify the bad action and construct fine-grained contrastive trajectory data.
2) This work primarily focuses on developing specialized LLM agents for a specific agent task, with limited exploration into the construction of strong generalized agents.
For future work, we will investigate the transferability of the policies trained by \method{} and try to apply our method in a multi-task training scenario.



\bibliography{custom}

\appendix

\clearpage
\onecolumn
\tcbset{width=\textwidth}

\section{Datasets}
\label{app:datasets}

\paragraph{WebShop} WebShop~\citep{yao2022webshop} is an online shopping website environment where agents navigate the platform to make purchases based on user instructions.
Once the agent selects the "buy" action, the environment provides a final reward, which is calculated based on the matching heuristics of the product's attributes and price.

\paragraph{ScienceWorld} ScienceWorld~\citep{wang2022scienceworld} is a text-based virtual environment centered around accomplishing elementary science experiments, including 10 different task types such as thermodynamics and electrical circuits.
The agents need to be grounded in embodied interactive environments to engage with and comprehend scientific concepts through practical experience.
Each task in ScienceWorld includes several optional subgoals, and the overall final reward is computed based on the achievement of these subgoals.

The original test set in ScienceWorld consists of critical unseen task variations.
For instance, in the training set, the task may involve boiling water, whereas in the test set, the task is to boil lead.
Consequently, we employ the original test set to evaluate our model's generalization performance on unseen scenarios.
We utilize the original development set as our test set with seen scenarios.
We exclude Task-9 and Task-10 due to their excessively long task-solving trajectories.
Following \citet{lin2023swiftsage}, we use the first 10 instances for task types with more than 10 test variations for fair and cost-effective comparisons.

\paragraph{ALFWorld} ALFWorld~\citep{shridhar2020alfworld} consists of interactive TextWorld environments that parallel embodied worlds in the ALFRED~\citep{shridhar2020alfred} dataset. 
In this environment, agents are required to explore and complete high-level house-holding instructions.
The original ALFWorld dataset comprises both seen and unseen evaluation sets.
The seen set is designed to assess in-distribution generalization, whereas the unseen set with new task instances measures out-of-distribution generalization of the agents.

\paragraph{CoT Annotation}
Webshop and ALFWorld provide a few human-annotated trajectories for imitation learning.
We also employ GPT-4 as the teacher agent to explore in the WebShop environment and select trajectories which have a reward greater than $0.7$.
ScienceWorld environment provides heuristic searching algorithms to generate golden trajectories for each sub-task.
Since the original trajectories do not contain CoT information for each action step, we use GPT-4 to generate the corresponding rationales.

\section{Success Rate}
\label{app:sr}

We report the success rate of our experiments in Table~\ref{tab:sr}.
Note the definition of success rates of three tasks are different.
For WebShop, success rate is defined as the portion of instances where final reward is 1.0.
For ScienceWorld, the original paper does not provide the definition of success rate. However, according to its official environment, a trajectory is considered success if the environment reaches a pre-defined latent state where the reward may not be exactly 1.0.
For ALFWorld, since it only provides binary final rewards, success rate is equal to average final reward.

\begin{table*}[t]
\vspace{-1em}
\centering
\tabcolsep=15pt
\resizebox{1\linewidth}{!}{
\begin{tabular}{@{}lccccc@{}}
\toprule
\multirow{2}{*}{\textbf{Method}} & \multirow{2}{*}{\textbf{WebShop}} & \multicolumn{2}{c}{\textbf{ScienceWorld}} & \multicolumn{2}{c}{\textbf{ALFWorld}} \\
\cmidrule(l){3-4} \cmidrule(l){5-6}
& & Seen & Unseen & Seen & Unseen \\
\midrule
GPT-4 & 34.0 & 58.3 & 59.7 & 42.9 & 38.1 \\
GPT-3.5-Turbo & 33.0 & 7.2 & 4.7 & 7.9 & 10.5 \\
\midrule
Llama-2-7B-Chat & 7.5 & 0.0 & 0.0 & 0.0 & 0.0 \\
Llama-2-7B-Chat + SFT & 33.0 & 70.6 & 73.5 & 60.0 & 67.2 \\
Llama-2-7B-Chat + Best-of-N & 35.0 & 70.6 & 73.5 & 62.1 & 69.4 \\
Llama-2-7B-Chat + RFT & 34.5 & 73.2 & 71.1 & 62.9 & 66.4 \\
Llama-2-7B-Chat + PPO & 33.5 & 68.5 & 70.5 & 22.1 & 29.1 \\
\midrule
Llama-2-7B-Chat + \method{} (ours) & \textbf{37.5} & \textbf{80.3} & \textbf{78.2} & \textbf{68.6} & \textbf{72.4} \\
\bottomrule
\end{tabular}
}
\caption{
The success rate of different methods on three agent datasets. For ALFWorld, success rate is equal to the average final reward.
}
\label{tab:sr}
\end{table*}

\section{Details for Step-Wise Contrastive}
\label{app:action}

We implement a variation of \method{} which learns from contrastive good-bad action pairs.
Specifically, for a task instruction $u$ with expert trajectory $e=(u,a_1,...,o_{n-1},a_n)$, we utilize teacher forcing for the first $t-1$ steps $(a_1,o_1,...a_{t-1},o_{t-1})$, and let the agent predict the actions from $t$-th step to get the trajectory:
\begin{equation}
    \hat{e}=(u,a_1,o_1,...,o_{t-1},\hat{a}_t,\hat{o}_t,...,\hat{o}_{m-1},\hat{a}_m)
\end{equation}
The environments return a reward $\hat{r}$ for the trajectory $\hat{e}$.
If we denote the golden trajectory for the first $t-1$ steps as $e_{(t-1)}$, then the good-bad action pairs $a_w\succ a_l\ \vert\ u,e_{(t-1)}$ is constructed based on the final rewards.
Here, $a_w$ and $a_l$ represent the actions with higher and lower final rewards, chosen from $(a_t,\hat{a}_t)$ respectively.
Then the contrastive relation of the action pair can also be utilized in DPO loss to improve the policy:
\begin{equation}
    \mathcal{L}_\mathrm{DPO}(\pi_\theta;\pi_\mathrm{ref}) = -\mathbb{E}\Bigg[\log \sigma \Big(\beta \log\frac{\pi_\theta(a_w\vert u,e_{(t-1)})}{\pi_\theta(a_l\vert u,e_{(t-1)})} - \beta\log\frac{\pi_\mathrm{ref}(a_w\vert u,e_{(t-1)})}{\pi_\mathrm{ref}(a_l\vert u,e_{(t-1)})}\Big)\Bigg].
\end{equation}

\section{Case Study}
\label{app:case}

Here, we present case studies to compare \method{} agents with SFT behavioral cloning agents.
We illustrate examples of WebShop and ScienceWorld in Figure~\ref{fig:webshop_case} and \ref{fig:sciworld_case}, respectively.

In the WebShop scenario, the SFT agent failed to select the ``3pc'' color option, resulting in a sub-optimal trajectory.
Conversely, \method{} demonstrated the ability to learn from past failures and adeptly chose the appropriate attribute option.

In the ScienceWorld example, the task is to find a(n) animal.
The SFT agent, however, persisted in performing the incorrect action, focusing on a non-living object instead.
Additionally, despite generating a correct CoT, the SFT agent's action selection was inconsistent with the CoT rationale.
In contrast, \method{} successfully accomplished the task, showcasing the effectiveness of trial and error.

\begin{tcolorbox}[breakable, title=Case Study of WebShop, enhanced jigsaw]
\columnseprule=0.5pt
\begin{multicols}{2}
\begin{center}\textcolor{red}{\textbf{\Large SFT Agent}}\end{center}
\textcolor{case_purple}{\textbf{Task Instruction:}} I am looking for a hair growth treatment in the color 3pc, and price lower than 40.00 dollars\\
\textcolor{case_blue}{\textbf{Thought 1:}} I should search for "hair growth treatment 3pc" first.\\
\textcolor{case_blue}{\textbf{Action 1:}} search[hair growth treatment 3pc]\\
\textcolor{case_purple}{\textbf{Observation 1:}} \textit{<searching results>}\\
\textcolor{case_blue}{\textbf{Thought 2:}} There are several options, but I need to find one that's within my budget. The first option, B09S12LVLZ, is a 3-piece ginger hair growth cream set priced at \$15.99, which is under \$50. This fits my criteria, so I'll take a closer look at this product.\\
\textcolor{case_blue}{\textbf{Action 2:}} click[B09S12LVLZ]\\
\textcolor{case_purple}{\textbf{Observation 2:}} \textit{<product details>}\\
\textcolor{case_blue}{\textbf{Thought 3:}} The product details confirm it's a 3-piece set and the price is still within my budget. I don't need to check the size options since it's already the 3-piece set I'm looking for. The price is good, and although there's no rating, I'm willing to try it out. I'll go ahead and make the purchase.\\
\textcolor{case_blue}{\textbf{Action 3:}} click[buy now]\\
\textcolor{case_purple}{\textbf{Reward:}} 0.5\\
\\
\sethlcolor{red!50}\hl{\textit{\# Did not select the color option.}}
\columnbreak

\begin{center}\textcolor{my_green}{\textbf{\Large ETO Agent}}\end{center}
\textcolor{case_purple}{\textbf{Task Instruction:}} I am looking for a hair growth treatment in the color 3pc, and price lower than 40.00 dollars\\
\textcolor{case_blue}{\textbf{Thought 1:}} I should search for "hair growth treatment 3pc" first.\\
\textcolor{case_blue}{\textbf{Action 1:}} search[hair growth treatment 3pc]\\
\textcolor{case_purple}{\textbf{Observation 1:}} \textit{<searching results>}\\
\textcolor{case_blue}{\textbf{Thought 2:}} B09S12LVLZ and B09QMJY2ZC are hair growth treatments in a 3pc set and priced lower than \$50. I can check B09S12LVLZ first.\\
\textcolor{case_blue}{\textbf{Action 2:}} click[B09S12LVLZ]\\
\textcolor{case_purple}{\textbf{Observation 2:}} \textit{<product details>}\\
\textcolor{case_blue}{\textbf{Thought 3:}} \sethlcolor{green!50}\hl{For a hair growth treatment in a 3pc set, the item has options for different sizes. I should click on the '3pcs' option.}\\
\textcolor{case_blue}{\textbf{Action 3:}} \sethlcolor{green!50}\hl{click[3pcs]}\\
\textcolor{case_purple}{\textbf{Observation 3:}} You have clicked [3pcs] option.\\
\textcolor{case_blue}{\textbf{Thought 4:}} Now that I have selected the 3pc option, I should proceed to purchase the item.\\
\textcolor{case_blue}{\textbf{Action 4:}} click[buy now]\\
\textcolor{case_purple}{\textbf{Reward:}} 1.0
\end{multicols}
\end{tcolorbox}
\begin{figure}[!h]
    \centering
    \vspace{-8pt}
    \caption{
    Case study for WebShop.
    }
    \label{fig:webshop_case}
\end{figure}

\begin{tcolorbox}[breakable,title=Case Study of ScienceWorld]
\columnseprule=0.5pt
\begin{multicols}{2}
\begin{center}\textcolor{red}{\textbf{\Large SFT Agent}}\end{center}

\textcolor{case_purple}{\textbf{Task Instruction:}} Your task is to find a(n) animal. First, focus on the thing. Then, move it to the yellow box in the bedroom.\\
\textcolor{case_blue}{\textbf{Thought 1:}} I need to find an animal as per the task description. I should start by looking around the current room to identify if there's an animal here.\\
\textcolor{case_blue}{\textbf{Action 1:}} look around\\
\textcolor{case_purple}{\textbf{Observation 1:}} This room is called the living room. In it, you see: the agent, a substance called air, a chair. On the chair is: nothing. a couch. On the couch is: a white pillow. a desk. On the desk is: a drawer. a finger painting. You also see: A door to the hallway (that is open).\\
\textcolor{case_blue}{\textbf{Thought 2:}} There are no animals in the living room, \sethlcolor{red!50}\hl{but I see a finger painting that could be considered an animal-related item}. I should look around again to ensure I didn't miss anything before deciding on my next step.\\
\textcolor{case_blue}{\textbf{Action 2:}} \sethlcolor{red!50}\hl{look around}\quad \sethlcolor{red!50}\hl{\textit{\# Repeat wrong action}}\\
\textcolor{case_purple}{\textbf{Observation 2:}} This room is called the living room. In it, you see: the agent, a substance called air, a chair. On the chair is: nothing. a couch. On the couch is: a white pillow. a desk. On the desk is: a drawer. a finger painting. You also see: A door to the hallway (that is open).\\
\textcolor{case_blue}{\textbf{Thought 3:}} The finger painting is a potential animal-related item, \sethlcolor{green!50}\hl{but it's not the target of the task. I should focus on finding an actual animal.}\\
\textcolor{case_blue}{\textbf{Action 3:}} \sethlcolor{red!50}\hl{focus on finger painting}\\
\sethlcolor{red!50}\hl{\textit{\# Not a(n) animal, contradict with the CoT}}\\
\textcolor{case_purple}{\textbf{Reward:}} 0.0\\
\\ \\ \\ \\ \\ \\ \\ \\ \\
\columnbreak

\begin{center}\textcolor{my_green}{\textbf{\Large ETO Agent}}\end{center}

\textcolor{case_purple}{\textbf{Task Instruction:}} Your task is to find a(n) animal. First, focus on the thing. Then, move it to the yellow box in the bedroom.\\
\textcolor{case_blue}{\textbf{Thought 1:}} I need to find an animal as per the task description. I should start by looking around the current location to identify any potential animals or related items.\\
\textcolor{case_blue}{\textbf{Action 1:}} look around\\
\textcolor{case_purple}{\textbf{Observation 1:}} This room is called the living room. In it, you see: ...\\
\textcolor{case_blue}{\textbf{Thought 2:}} The living room doesn't seem to have any animals, but I might have missed something. I should check another area.\\
\textcolor{case_blue}{\textbf{Action 2:}} teleport to outside\\
\textcolor{case_purple}{\textbf{Observation 2:}} You teleport to the outside.\\
\textcolor{case_blue}{\textbf{Thought 3:}} Now that I'm outside, I should look around to see if there are any animals or indications of animals here.\\
\textcolor{case_blue}{\textbf{Action 3:}} look around\\
\textcolor{case_purple}{\textbf{Observation 3:}} This outside location is called the outside. Here you see: the agent, an axe, a frog egg, a baby wolf ...\\
\textcolor{case_blue}{\textbf{Thought 4:}} I see a frog egg here, which is a developmental stage of an animal. This fits the task description of finding an animal.\\
\textcolor{case_blue}{\textbf{Action 4:}} \sethlcolor{green!50}\hl{focus on frog egg}\\
\textcolor{case_purple}{\textbf{Observation 4:}} You focus on the frog egg.\\
\textcolor{case_blue}{\textbf{Thought 5:}} The frog egg is the early stage of a common toad, which is an animal. I should pick it up to move it to the specified location.\\
\textcolor{case_blue}{\textbf{Action 5:}} pick up frog egg\\
\textcolor{case_purple}{\textbf{Observation 5:}} You move the common toad to the inventory.\\
\textcolor{case_blue}{\textbf{Thought 6:}} With the common toad (frog egg) in my inventory, I need to go to the bedroom to place it in the yellow box as instructed.\\
\textcolor{case_blue}{\textbf{Action 6:}} teleport to bedroom\\
\textcolor{case_purple}{\textbf{Observation 6:}} You teleport to the bedroom.\\
\textcolor{case_blue}{\textbf{Thought 7:}} Now that I'm in the bedroom, I should complete the task by moving the common toad from my inventory to the yellow box.\\
\textcolor{case_blue}{\textbf{Action 7:}} \sethlcolor{green!50}\hl{move common toad to yellow box}\\
\textcolor{case_purple}{\textbf{Reward:}} 1.0
\end{multicols}
\end{tcolorbox}
\begin{figure}[ht]
    \centering
    \vspace{-8pt}
    \caption{
    Case study for ScienceWorld.
    }
    \label{fig:sciworld_case}
\end{figure}

\section{Prompt for Evaluation}
\label{app:prompt}

We show the instruction prompts for WebShop, ScienceWorld, ALFWorld in Figure~\ref{fig:webshop_prompt}, \ref{fig:sciworld_prompt}, \ref{fig:alfworld_prompt}, respectively.

\begin{tcolorbox}[breakable,title=Instruction Prompt for WebShop ]
You are doing a web shopping task.
I will give you instructions about what to do.
You have to follow the instructions. Every round I will give you an observation and a list of available actions, you have to respond to an action based on the state and instruction.
You can use search action if search is available.
You can click one of the buttons in clickables.
An action should be one of the following structure: search[keywords] or click[value]\\

If the action is not valid, perform nothing.
Keywords in search are up to you, but the value in click must be a value in the list of available actions.
Remember that your keywords in search should be carefully designed.\\

Your response should use the following format:\\
Thought: I think ...\\
Action: click[something]
\end{tcolorbox}
\begin{figure}[ht]
    \centering
    \vspace{-8pt}
    \caption{
    Instruction prompt for WebShop.
    }
    \label{fig:webshop_prompt}
\end{figure}

\begin{tcolorbox}[breakable, title=Instruction Prompt for ScienceWorld, enhanced jigsaw]
You are a helpful assistant to do some scientific experiments in an environment.
In the environment, there are several rooms: kitchen, foundry, workshop, bathroom, outside, living room, bedroom, greenhouse, art studio, hallway
You should explore the environment and find the items you need to complete the experiment.
You can teleport to any room in one step.
All containers in the environment have already been opened, you can directly get items from the containers.\\

The available actions are:\\
open OBJ: open a container\\
close OBJ: close a container\\
activate OBJ: activate a device\\
deactivate OBJ: deactivate a device\\
connect OBJ to OBJ: connect electrical components\\
disconnect OBJ: disconnect electrical components\\
use OBJ [on OBJ]: use a device/item\\
look around: describe the current room\\
examine OBJ: describe an object in detail\\
look at OBJ: describe a container's contents\\
read OBJ: read a note or book\\
move OBJ to OBJ: move an object to a container\\
pick up OBJ: move an object to the inventory\\
pour OBJ into OBJ: pour a liquid into a container\\
mix OBJ: chemically mix a container\\
teleport to LOC: teleport to a specific room\\
focus on OBJ: signal intent on a task object\\
wait: task no action for 10 steps\\
wait1: task no action for a step\\

Your response should use the following format:\\
Thought: <your thoughts>\\
Action: <your next action>
\end{tcolorbox}
\begin{figure}[ht]
    \centering
    \vspace{-8pt}
    \caption{
    Instruction prompt for ScienceWorld.
    }
    \label{fig:sciworld_prompt}
\end{figure}

\begin{tcolorbox}[breakable, title=Instruction Prompt for ALFWorld, enhanced jigsaw]
Interact with a household to solve a task. Imagine you are an intelligent agent in a household environment and your target is to perform actions to complete the task goal. At the beginning of your interactions, you will be given a detailed description of the current environment and your goal to accomplish. \\
For each of your turn, you will be given the observation of the last turn. You should first think about the current condition and plan for your future actions, and then output your action in this turn. Your output must strictly follow this format:"Thought: your thoughts. Action: your next action".\\

The available actions are:\\
1. go to {recep}\\
2. task {obj} from {recep}\\
3. put {obj} in/on {recep}\\
4. open {recep}\\
5. close {recep}\\
6. toggle {obj} {recep}\\
7. clean {obj} with {recep}\\
8. heat {obj} with {recep}\\
9. cool {obj} with {recep}\\
where {obj} and {recep} correspond to objects and receptacles.\\
After each turn, the environment will give you immediate feedback based on which you plan your next few steps. if the environment outputs "Nothing happened", that means the previous action is invalid and you should try more options.\\

Your response should use the following format:\\
Thought: <your thoughts>\\
Action: <your next action>
\end{tcolorbox}
\begin{figure}[ht]
    \centering
    \vspace{-8pt}
    \caption{
    Instruction prompt for ALFWorld.
    }
    \label{fig:alfworld_prompt}
\end{figure}

\end{document}